\pdfoutput=1

\documentclass[11pt]{article}

\usepackage{ACL2023}

\usepackage{times}
\usepackage{latexsym}

\usepackage[T1]{fontenc}

\usepackage[utf8]{inputenc}

\usepackage{microtype}

\usepackage{inconsolata}

\usepackage{graphicx}
\usepackage{booktabs}
\usepackage{algorithm}
\usepackage[noend]{algpseudocode}
\usepackage{amsmath}
\usepackage{multirow}
\usepackage{subfigure}
\usepackage{colortbl} 
\DeclareMathOperator*{\argmax}{arg\,max}
\DeclareMathOperator*{\argmin}{arg\,min}
\DeclareMathOperator*{\sample}{\text{sample}}
\usepackage{cleveref}
\usepackage{anyfontsize}

\expandafter\def\expandafter\normalsize\expandafter{%
    \normalsize%
    \setlength\abovedisplayskip{4pt}%
    \setlength\belowdisplayskip{4pt}%
    \setlength\abovedisplayshortskip{-8pt}%
    \setlength\belowdisplayshortskip{0pt}%
}
\title{Draft \& Verify: Lossless Large Language Model Acceleration via Self-Speculative Decoding}

\author{Jun Zhang$^1$, Jue Wang$^1$, Huan Li$^1$, Lidan Shou$^1$, \\ \textbf{Ke Chen$^1$, Gang Chen$^1$, Sharad Mehrotra$^2$} \\
  $^1$The State Key Laboratory of Blockchain and Data Security, Zhejiang University \\
  $^2$Donald Bren School of Information and Computer Sciences, University of California, Irvine \\
  \texttt{\{zj.cs, zjuwangjue, lihuan.cs, should, chenk, cg\}@zju.edu.cn},\\ \texttt{sharad@ics.uci.edu} 
}

\begin{document}
\maketitle
\begin{abstract}

We present a novel inference scheme, self-speculative decoding, for accelerating Large Language Models (LLMs) without the need for an auxiliary model. 
This approach is characterized by a two-stage process: drafting and verification.
The drafting stage generates draft tokens at a slightly lower quality but more quickly,
which is achieved by selectively skipping certain intermediate layers during drafting.
Subsequently, the verification stage employs the original LLM to validate those draft output tokens in one forward pass.
This process ensures the final output remains \textit{identical} to that produced by the unaltered LLM.
Moreover, the proposed method requires no additional neural network training and no extra memory footprint, 
making it a plug-and-play and cost-effective solution for inference acceleration.
Benchmarks with LLaMA-2 and its variants demonstrated a speedup up to 1.99$\times$.\footnote{Code is available at \url{https://openreview.net/attachment?id=ACC2nQYzPYS&name=software}, and will be released with the Apache-2.0 License.}

\end{abstract}

\section{Introduction}

Transformer-based Large Language Models (LLMs), such as GPT-3/4, PaLM, and LLaMA, have been widely adopted in various real-world applications \cite{bommasani2021opportunities,liang2022holistic,brown2020language,min2022rethinking,chan2022data,touvron2023llama}. 
However, their inference costs have raised significant concerns, especially for latency-sensitive scenarios \cite{pope2022efficiently}. 
The main efficiency bottleneck is the \textit{autoregressive decoding} process. 
This process decodes each output token sequentially, leading to a high number of Transformer calls; 
furthermore, each Transformer call is typically memory bandwidth-bound, 
resulting in low computation utility and thus longer wall-clock time \cite{shazeer2019fast}.
For instance, decoding 128 tokens autoregressively using LLaMA-2-13B on an A100 GPU can take up to 100$\times$ longer than a sequence-level forward pass on the same number of tokens, highlighting the substantial inefficiency inherent in the current decoding process.

Established model compression techniques such as 
quantization \cite{han2015deep}, 
pruning \cite{molchanov2016pruning}, 
and distillation \cite{hinton2015distilling} have been employed to alleviate these costs. 
While these solutions have proven extremely effective, 
they usually require changing the model architecture, 
changing the training procedure, re-training or fine-tuning the models, 
and do not maintain identical outputs.

In parallel to model compression, 
\textit{speculative execution} is being explored to accelerate the autoregressive decoding process
\cite{leviathan2023fast,chen2023accelerating}.
These methods train an auxiliary \textbf{draft model} that can quickly generate some draft output tokens. 
Subsequently, the original LLM, referred to as the \textbf{verify model}, then checks the acceptability of these draft tokens with one single forward pass.
This verification step ensures that the outputs are derived from the original LLM's probability distribution.

However, an essential issue of existing speculative execution methods is the need to identify or train a suitable draft model that can generate outputs consistent with the verify model.
It becomes more tricky when the LLM is already a fine-tuned model, e.g.~LLaMA-2-Chat \cite{touvron2023llama}, CodeLLaMA \cite{Rozire2023CodeLO}. 
How to find or train a draft model that can effectively mimic the outputs of such a tailored model is a formidable task, with no straightforward or guaranteed solutions.
Furthermore, the introduction of an additional draft model escalates the GPU memory overhead, increasing deployment challenges particularly on devices with restricted memory capacity.


\begin{figure*}
    \centering
    \includegraphics[width=\linewidth]{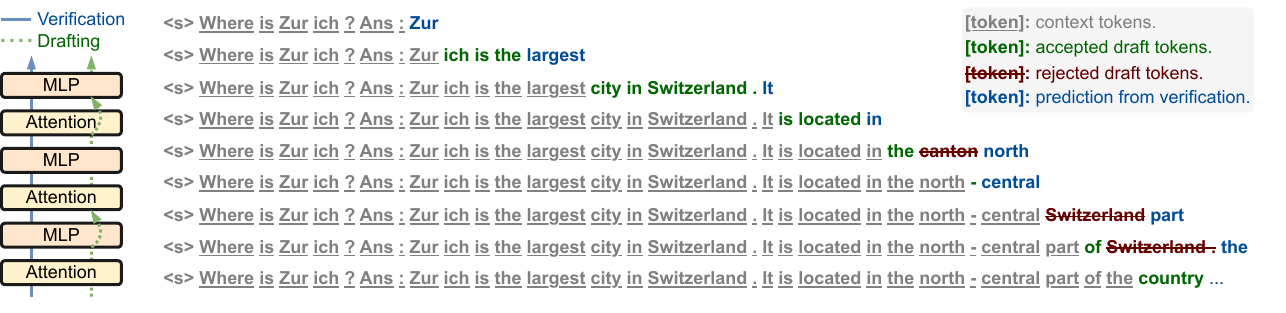}
    \caption{Visualization of the self-speculative decoding process. 
            The verification stage evaluates all drafted tokens in a single forward pass, with accepted tokens marked in green and rejected tokens highlighted in red. Each verification step also predicts one more token, which is denoted in blue. }
    \label{fig:intro}
\end{figure*}

In this paper, we present \textit{self-speculative decoding}, a novel approach to accelerate the inference of LLMs. 
This method builds on the principles of speculative execution, but with a unique twist: 
it utilizes one LLM for both drafting and verification stages.
The key insight driving our approach is the observation that 
skipping certain layers in LLMs does not significantly compromise the generation quality \cite{liu2023deja}.
As such, by selectively bypassing some intermediate layers, we can use the LLM itself to generate draft tokens. 
These tokens are then verified by the original LLM in a single forward pass.
\Cref{fig:intro} illustrates this two-stage decoding process.
The blue arrow indicates the inference path of the original model, while the green arrow depicts the inference path during the drafting stage.
Notably, both inference paths share the same model so we do not need a standalone draft model with extra memory overhead.

\begin{figure}
    \centering
    \includegraphics[width=0.9\linewidth]{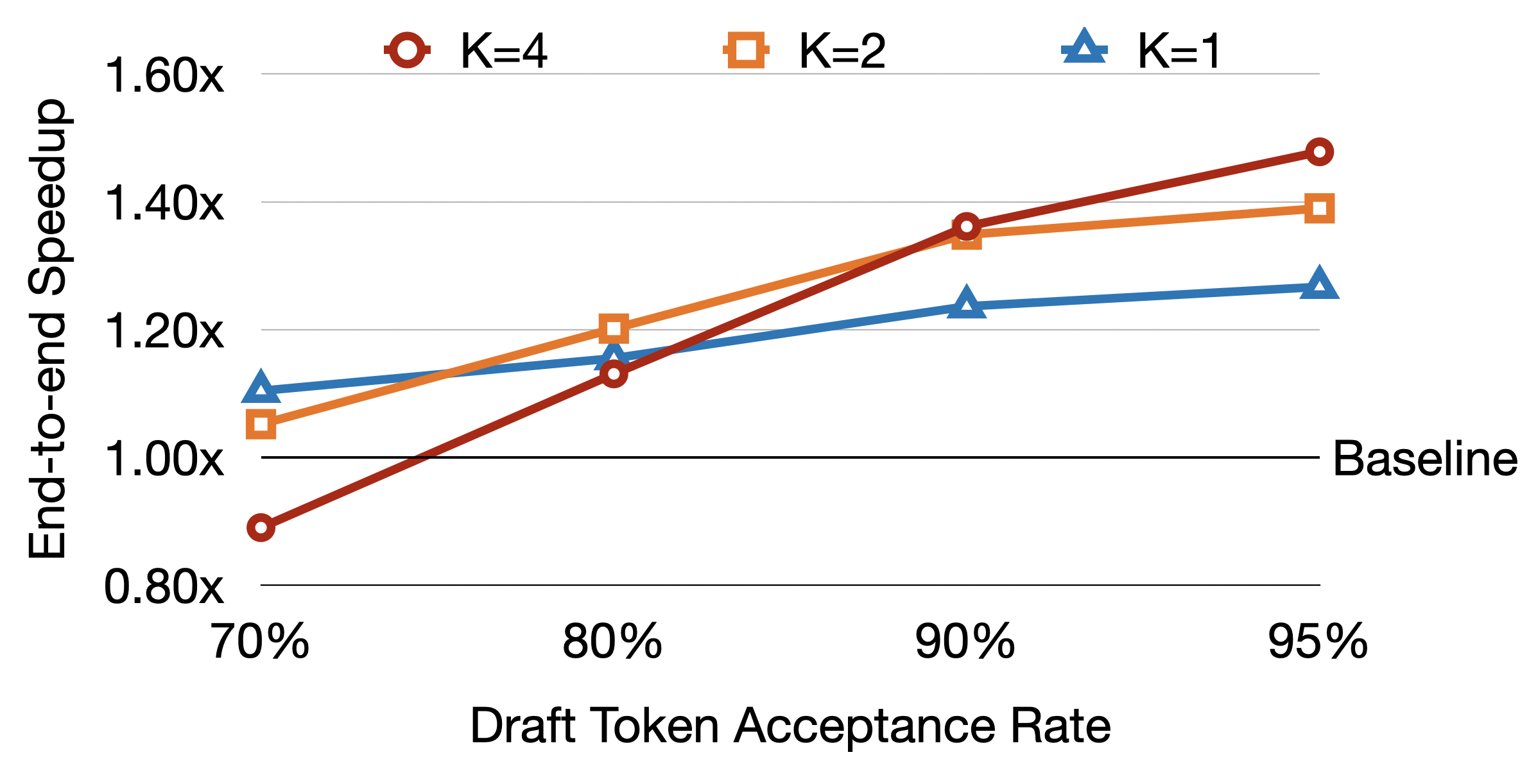}
    \caption{Impact of the number of draft tokens ($K$) and acceptance rate on end-to-end speedup. The draft model is assumed to be 2$\times$ faster than the verify model.}
    \label{fig:analyse}
\end{figure}

Implementing self-speculative decoding poses two main challenges: 
(a) determining which layers and the number of layers to skip during drafting, 
and (b) deciding the timing to stop generating draft tokens.
To tackle the first challenge, we formulate it as an optimization problem, 
which accepts the combinations of layers to bypass as input and aims to minimize the average inference time per token. 
We employ Bayesian optimization \cite{jones1998efficient} to solve this problem. 
The optimization is performed offline at the model level, and the searched layer combinations are fixed.
The second challenge pertains to determining the optimal number of draft tokens ($K$) to generate. 
As shown in \Cref{fig:analyse}, the choice of $K$ significantly influences the end-to-end speedup:
for an acceptance rate below 80\%, $K=1$ is optimal, and for rates above 80\%, a larger $K$ is necessary. 
This observation underscores that a static $K$ is not universally applicable. 
To tackle this variability, we introduce an adaptive draft-exiting mechanism, 
which stops generating draft tokens once its confidence level drops below a threshold. 
This intervention prevents unnecessary computation and potential discard of additional draft tokens, thereby enhancing efficiency.

To summarize, our main contributions are:
(1) \textit{Inference scheme}: we propose self-speculative decoding, a practical, plug-and-play solution for inference acceleration that does not require further neural network training and avoids additional memory overhead;
(2) \textit{Optimization strategies}: we adopt Bayesian optimization to select which layers to skip during drafting and propose a simple yet effective method to adaptively determine the number of draft tokens;
(3) \textit{Evaluation}: we evaluate our method on text summarization and code generation tasks,
and the experimental results indicate that our method can achieve up to 1.99$\times$ in end-to-end speedup.

\section{Related Work}

\paragraph{Transformer-based LLM inference.} 
As LLMs continue to evolve rapidly, we are seeing a surge of systems specifically engineered for LLM inference, including Faster Transformer \cite{nvidiaft},  Orca \cite{yu2022orca}, LightSeq \cite{wang2021lightseq}, PaLM inference \cite{pope2022efficiently}, TurboTransformers \cite{fang2021turbotransformers}, Deepspeed Inference~\cite{aminabadi2022deepspeed},
FlexGen~\cite{sheng2023high},
Text Generation Inference \cite{TGI}, etc. 
The token generation phase typically takes up the majority of the end-to-end inference time compared to the prompting encoding phase.
Despite the introduction of system optimizations by those state-of-the-art systems to improve the inference speed, there is still a gap in the careful co-design of algorithms and systems. This is necessary to fully harness the potential of hardware efficiency during LLM inference computation.

\paragraph{Model Compression.} 
Various model compression methods have been studied for model inference. 
For example, quantization~\cite{han2015deep, jacob2018quantization,nagel2019data,zhao2019improving,yao2022zeroquant,park2022nuqmm,dettmers2022llm,xiao2022smoothquant,frantar2022gptq}, pruning or sparsification~\cite{molchanov2016pruning,liu2018rethinking,he2019filter,hoefler2021sparsity,frantar2023massive,liu2023deja,bansal2022rethinking}, and distillation~\cite{hinton2015distilling,cho2019efficacy,tang2019distilling,touvron2021training}  have been applied to speed up the inference of the machine learning model, particularly LLMs.
While these solutions are extremely effective, 
they often necessitate modifications to the model architecture and the training procedure. 
This usually involves re-training or fine-tuning the models.
And it is important to note that these methods do not result in identical outputs.


\paragraph{Speculative Execution.} 
Speculative execution \cite{burton1985speculative,hennessy2011computer} is employed in computer architecture where a system performs
some task in advance if that task is known to be required after the previous task.
Speculative decoding \cite{chen2023accelerating,leviathan2023fast} has been proposed as an effective strategy to boost the inference speed of LLMs. 
Previously, \cite{stern2018blockwise} proposed to use block-wise parallel decoding to accelerate greedy decoding of attention models.
However, these methods need to train or select a high-quality draft model, and also result in increased memory overhead.
\citet{yang2023inference} proposed to copy the reference text tokens and validate them in a forward pass. 
However, this method relies on the repetitiveness assumption, and thus does not apply to general scenario generation.
In contrast, our approach does not incur additional memory overhead and does not hinge on explicit assumptions about data distribution.

\paragraph{Early Exit.} 
Early exit allows the model to choose different calculation paths based on the input during the inference process to achieve acceleration.
Various early exit techniques for encoder-only Transformers \cite{DevlinCLT19} have been proposed \cite{xin2020deebert,SchwartzSSDS20,LiuZWZDJ20,xin20early,hou20dynabert,zhou20bert,liao2021global,zhu21leebert,LiSSYQH20,SunLZGWHNXHQ22}.
Recently, \cite{schuster2022confident} further verified the effectiveness of early exit on the encoder-decoder LLM \cite{2020t5}.
Inspired by these works, we opt to skip certain intermediate layers during drafting.

\section{Method}

In this section, we first go through the standard autoregressive decoding. 
Subsequently, we provide a detailed depiction of our proposed method,
including selectively skipping layers during drafting, 
and adaptively determining the number of draft tokens.

\subsection{Standard Autoregressive Decoding}

\begin{algorithm}[t]
\footnotesize
\caption{Autoregressive Decoding (Greedy)}\label{alg:ARD}
\begin{algorithmic}[1]
\State Given model $p(x|x_1, ..., x_t)$, prompt $x_1, ..., x_t$ and target sequence length $T$.
\For{i = t, ..., T-1}
\State $x_{i+1} \leftarrow \argmax{p(x|x_1, ..., x_{i})}$
\EndFor
\State \textbf{return} $x_1, ..., x_T$
\end{algorithmic}
\end{algorithm}

Existing LLMs typically follow an autoregressive decoding process.
Given a prompt sequence $x_1, ..., x_t$, the model calculates the probability distribution of the next token $p(x|x_1, ..., x_{t})$. 
We present a greedy decoding process in \Cref{alg:ARD}.
In practice, instead of choosing the token with the highest probability (as in greedy decoding), we can sample tokens based on their probability distribution,
which introduces some randomness and generates more diverse outputs.

Ideally, the computational cost of autoregressive decoding is comparable to that of sequence-level forward processing for an equivalent number of tokens.\footnote{
In fact, due to the causal nature of language modeling, autoregressive decoding could potentially save some attention computation.
} 
However, this decoding process is significantly bounded by the memory bandwidth of the device.
When decoding each token, all the model parameters need to pass through the accelerator chip.
So the model size divided by the memory bandwidth gives a hard ceiling 
on the decoding speed, resulting in a much longer inference time.



\subsection{Self-Speculative Decoding}

To mitigate the inherent inefficiency of autoregressive decoding, speculative decoding can be employed to enhance the inference speed of LLMs. 
This strategy involves two models: an LLM that we want to optimize, and a draft model that runs faster, albeit potentially at a lower quality.
Speculative decoding can be explained as a two-stage process:
(1) drafting: the draft model first generates $K$ draft tokens from a given prompt sequence $x_1, ..., x_i$, denoted as $x_{i+1}, ..., x_{i+K}$.
(2) verification: following the drafting stage, the original LLM is then employed to validate these draft tokens. This validation is accomplished in a single forward pass, where the LLM predicts the probability distributions for each draft token and assesses whether they align with the draft tokens. 
Once a draft token $x_{j}$ is not validated, 
we use the original LLM's prediction to override $x_{j}$, and start the next round of drafting beginning from token $x_{j+1}$.

The above process is based on the observation that computing the forward pass of a short continuation of tokens in parallel is not much slower than that of a single token.
Consequently, the verification stage could be significantly more efficient than decoding tokens using the original LLM in standard autoregressive decoding.

\begin{algorithm}[t]
\footnotesize
\caption{Self-Speculative Decoding (Greedy)}\label{alg:SSD}
\begin{algorithmic}[1]
\State LLM $p(x|\boldsymbol z^*,x_1, ..., x_t)$ where $x_1, ..., x_t$ is the prompt, $\boldsymbol z^*$ is a vector that represents the specific layers to bypass;
target sequence length $T$;
max draft tokens to generate $K$.
The $p(x|\vec{0},x_1, ..., x_t)$ denotes original LLM,
where $\vec{0}$ is a zero vector, indicating all layers are used in inference.
\State $i \leftarrow t$
\While{$i < T$}
\For{$j \leftarrow i, ..., i + K$} \Comment{Drafting Stage}
    \State $x_{j+1} \leftarrow \argmax{p(x|\boldsymbol z^*,x_1, ..., x_{j})}$    
    \If{need to exit drafting (\S\ref{sec:exit})}
        \State Break
    \EndIf
\EndFor
\For{$i \leftarrow i, ..., j$} \Comment{Verification Stage}
    \If{$x_{i+1} \ne \argmax{p(x|\vec{0}, x_1, ..., x_{i})}$}
        \State $x_{i+1} \leftarrow \argmax{p(x|\vec{0}, x_1, ..., x_{i})}$ 
        \State Break
    \EndIf
\EndFor
\State $i \leftarrow i + 1$
\State If all draft tokens are accepted, 
        generate next token $x_{i+1} \leftarrow \argmax{p(x|\vec{0},x_1, ..., x_{i})}$ and $i \leftarrow i+1$
\EndWhile
\State \textbf{return} $x_1, ..., x_T$
\end{algorithmic}
\end{algorithm}

In contrast to existing methods that use a standalone draft model to obtain draft tokens, 
our paper proposes a novel `self-speculative' approach.
We employ the original LLM itself for both the drafting and verification stages. 
During the drafting stage, the LLM selectively skips some of its intermediate layers so as to generate draft tokens quicker.
Subsequently, these draft tokens are verified by the original LLM. 
\Cref{alg:SSD} presents a detailed description of the greedy decoding process.
A complete sampling-based decoding process is elaborated in \Cref{apd:samping_algo}.

Despite the simplicity of the main idea of self-speculative decoding,
it poses several challenges:

\textbf{Challenge 1:}
First, it is non-trivial to determine which layers and the number of layers to skip during drafting.
If an excessive number of layers are skipped, the quality of the draft could be significantly compromised. This could result in a low acceptance rate in the verification stage, consequently increasing the overall inference time. 
On the other hand, if fewer layers are skipped, it ensures a higher acceptance, but also caps the maximum speedup that could be achieved.

\textbf{Challenge 2:}
It is hard to decide when to stop the generation of draft tokens. 
As shown in \Cref{fig:analyse}, the choice of the number of draft tokens to generate significantly influences the end-to-end speedup.
In speculative decoding, if a draft token is rejected, all subsequent draft tokens will be discarded. 
Therefore, generating an excessive number of draft tokens could lead to unnecessary computation, thereby increasing the inference time.

In \Cref{sec:selection,sec:exit}, we will detail our approach to address these two challenges respectively.

\subsection{Selection of Skipped Layers}  \label{sec:selection}

The selection of skipped layers is essential in our method, shaping the configuration of the draft model and, consequently, the speedup achieved via self-speculative decoding. This selection process must carefully balance two key factors: the draft model's `effectiveness' and `efficiency.' Both are intrinsically linked to the selection of skipped layers and significantly influence the overall performance of our method. Specifically:

(1) The `effectiveness' of the draft model is quantified by the acceptance rate, which measures the agreement between the draft and verify models. 
However, we note that an exclusive focus on optimizing the acceptance rate could lead to no layers being skipped, i.e.~the draft model identical to the verify model, resulting in an acceptance rate of 100\%, but without any speedup.

(2) On the other hand, the `efficiency' of the draft model can be quantified by the number of layers in the draft model. 
Indeed, minimizing the number of layers can reduce the inference latency of the draft model, but an extreme setup where all layers are skipped would result in the draft model generating low-quality tokens. This would drastically lower the acceptance rate, negating any potential speedup.


In this section, we frame the layer selection process as an optimization problem, our primary objective is to optimize the \textit{average inference time per verified token}. This metric provides a comprehensive measure of the end-to-end inference time, including both drafting and verifying stages, normalized by the number of verified tokens.


\textbf{Objective Function.}
This metric is a function of the selection of layers to be skipped in the draft model. The function, represented as $f(\boldsymbol z)$, takes the selection of layers ($\boldsymbol z$) as input and returns the average inference time per \textit{verified} token on a development set. Here, $\boldsymbol z$ is a vector that represents the layers to be skipped.



\textbf{Optimization Problem.}
The optimization problem's goal is to find the input $\boldsymbol z^*$ that minimizes the objective function $f(\boldsymbol z)$. This problem can be formally expressed as:
\begin{align}
\boldsymbol z^* = \argmin_{\boldsymbol z} f(\boldsymbol z), \quad s.t.\ \boldsymbol z \in \{0, 1\}^{L}.
\end{align}

While a brute force search could find the globally optimal solution for smaller models with a manageable solving space, it becomes prohibitively expensive for larger language models with many layers (e.g., $L=160$ for LLaMA-2-70B).

\begin{figure}
    \centering
    \includegraphics[width=\linewidth]{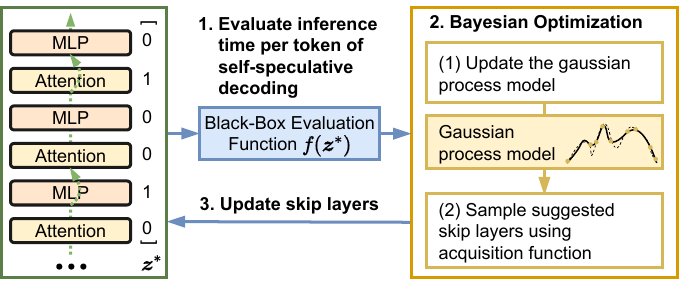}
    \caption{Illustration of using Bayesian optimization to search the best combination of skip layers that results in the lowest average token inference time.}
    \label{fig:bo}
\end{figure}

To tackle this, we employ Bayesian optimization \cite{jones1998efficient}. 
As shown in \Cref{fig:bo}, it iteratively selects new inputs $\boldsymbol z^*$ for evaluation, based on a surrogate model of the objective function, i.e.~Gaussian process \cite{rasmussen2006gaussian}, and an acquisition function. The latter balances exploration (testing inputs where the model's prediction is uncertain) and exploitation (testing inputs where the model anticipates a favorable result). 
This procedure continues until a predetermined number of iterations is reached.
We use the obtained $\boldsymbol z^*$ to accelerate text generation, and $\boldsymbol z^*$ is fixed for each model (i.e.,~generating draft model at model-level) without further updating.
When the draft model's target tasks vary significantly, task-level optimization is more appropriate to achieve good performance. Specifically, building the development set of optimization process from a single data source to mitigate inter-task interference.

In addition, while we here adopt skipping intermediate layers as a simple yet effective strategy to expedite the drafting stage, our scheme can be integrated with quantization \cite{dettmers2022llm} and sparsification \cite{sun2023wanda} to further reduce resource consumption, as detailed in \Cref{apd:adaption}.

\subsection{Adaptive Draft-Exiting Mechanism} \label{sec:exit}
Our self-speculative decoding approach incorporates an adaptive draft-exiting mechanism to enhance computational efficiency during the drafting stage. 
In speculative decoding, if a draft token is rejected, all subsequent draft tokens will be discarded accordingly.
The draft-exiting mechanism prevents the wasteful allocation of computational resources toward draft tokens that are less likely to be accepted in the verification stage.

Specifically, it compares the predicted probability of each draft token against a threshold $\gamma$. 
If the predicted probability falls below $\gamma$ such that $p(x_{t+1}|x_1,...,x_t) < \gamma$, 
indicating low confidence, it immediately stops drafting.
This approach ensures a better use of computing by focusing on the generation and verification of high-quality tokens, thereby improving the overall efficiency.

Moreover, it is worth noting that a static threshold may not accurately reflect the actual acceptance rate between the drafting and verification stages. 
For example, more challenging examples with a lower acceptance rate would be better served by a higher $\gamma$.
To avoid the need for case-by-case threshold determination, we use an adaptive threshold that adjusts dynamically according to an updating rule, 
thereby allowing for an accurate reflection of the acceptance rate and better handling of examples in different difficulties. 
We denote the acceptance rate ($AR$) at $e$-th drafting stage as $AR_e$. 
Consequently, the update rule is defined as follows:
\begin{align}
AR & \gets \beta_{1} AR + (1 - \beta_{1})AR_e, \\
\tilde{\gamma} &= 
\begin{cases}
    \gamma + \epsilon, & \text{if} \ {AR} \leq \alpha \\ 
    \gamma - \epsilon, & \text{otherwise}
\end{cases}, \\
\gamma & \gets \beta_{2} \gamma + (1 - \beta_{2}) \tilde{\gamma},
\end{align}
where
$\alpha$ represents a target acceptance rate, 
$\epsilon$ is the update step-size, and $\beta_{1}$ and $\beta_{2}$ are factors designed to mitigate fluctuations of $\gamma$ and $AR$ respectively. Notably, when $e$ is 1, $\beta_{1}=0$.
We update $\gamma$ after each verification stage.
This updating rule ensures that the acceptance rate remains in close proximity to a target acceptance rate $\alpha$.

\section{Evaluation}

\subsection{Setup}

We evaluate a diverse range of models including
LLaMA-2-13B, LLaMA-2-13B-Chat, CodeLLaMA-13B, and LLaMA-2-70B. 
Detailed setup can be found in \Cref{apd:setup}.
We perform Bayesian optimization\footnote{\url{https://github.com/bayesian-optimization/BayesianOptimization} (MIT License) is used.} (BO) for 1000 iterations to select the skipped layers in the drafting stage\footnote{\Cref{apd:setup} reports the offline BO time at model-level.}.
Results of tuning the number of BO iterations are reported in \Cref{apd:boiter}.
The datasets include CNN/Daily Mail (CNN/DM), Extreme Summarization (XSum), and HumanEval. 
These tasks cover the evaluation of text and code generation capabilities.
\Cref{apd:diverse_task} shows effectiveness on more diverse tasks such as solving math problems and open-domain chitchat.
We perform 1-shot evaluation for CNN/DM and XSum, and compare the ROUGE-2 \cite{lin2004rouge}.
We compare pass@1 and pass@10 \cite{kulal2019spoc} for HumanEval.
We randomly sample 1000 instances from the testset for CNN/DM and XSum.





\subsection{Main Results}
\begin{table*}[!t]
    \centering
    \footnotesize
    \renewcommand{\arraystretch}{0.9}
    \setlength{\tabcolsep}{1.4mm}
    \begin{tabular}{llccrcr}
    \toprule
    \multirow{2}{*}{Model}
    & \multirow{2}{*}{Method}
    & \multirow{2}{*}{Temp.}
    & \multicolumn{2}{c}{CNN/DM}  & \multicolumn{2}{c}{XSum}          \\ \cmidrule(l){4-5} \cmidrule(l){6-7} &    &     & \multicolumn{1}{l}{ROUGE-2} & Speedup   & \multicolumn{1}{l}{ROUGE-2} & Speedup   \\ \midrule
    LLaMA-2-13B      & Autoregressive     & 0.0         & 0.106 & 1.000$\times$  & 0.124  & 1.000$\times$      \\
    LLaMA-2-13B      & Self-Speculative   & 0.0         & 0.108 & 1.572$\times$  & 0.125  & 1.429$\times$ \\
    LLaMA-2-13B      & Autoregressive     & 0.2         & 0.111 & 1.000$\times$  & 0.117  & 1.000$\times$      \\
    LLaMA-2-13B      & Self-Speculative   & 0.2         & 0.111 & 1.529$\times$  & 0.117  & 1.377$\times$ \\ \midrule
    LLaMA-2-13B-Chat & Autoregressive     & 0.0         & 0.144 & 1.000$\times$  & 0.109  & 1.000$\times$      \\
    LLaMA-2-13B-Chat & Self-Speculative   & 0.0         & 0.143 & 1.409$\times$  & 0.109  & 1.224$\times$ \\
    LLaMA-2-13B-Chat & Autoregressive     & 0.2         & 0.143 & 1.000$\times$  & 0.106  & 1.000$\times$      \\
    LLaMA-2-13B-Chat & Self-Speculative   & 0.2         & 0.145 & 1.383$\times$  & 0.108  & 1.210$\times$        \\ \midrule
    LLaMA-2-70B      & Autoregressive     & 0.0         & 0.130 & 1.000$\times$  & 0.118   & 1.000$\times$      \\
    LLaMA-2-70B      & Self-Speculative   & 0.0         & 0.130 & 1.992$\times$ 
                                                        & 0.118 & 1.598$\times$ \\ 
    LLaMA-2-70B      & Autoregressive     & 0.2         & 0.131 & 1.000$\times$      &  0.108   & 1.000$\times$      \\
    LLaMA-2-70B      & Self-Speculative   & 0.2         & 0.131 & 1.964$\times$ 
                                                        & 0.110 & 1.560$\times$ \\ 
    \bottomrule
    \end{tabular}
    \caption{Evaluation on text generation tasks. `Speedup' represents the acceleration of average inference time per token compared to the `Autoregressive' baseline on the same setting.}
    \label{tab:text}
\end{table*}

\begin{table}[!t]
    \centering
    \footnotesize
    \renewcommand{\arraystretch}{1.0}
    \setlength{\tabcolsep}{1.4mm}
    \begin{tabular}{@{}lllcr@{}}
    \toprule
    Model       &   Method & \multicolumn{2}{c}{HumanEval} & Speedup \\ 
    \midrule
    {CodeLLaMA-13B}   &  Autoreg.   & {pass@1}         & 0.311         & 1.000$\times$   \\
    {CodeLLaMA-13B}   &  Self-Spec. & {pass@1}         & 0.317         & 1.456$\times$   \\
    {CodeLLaMA-13B}   &  Autoreg.   & {pass@10}        & 0.659         & 1.000$\times$   \\
    {CodeLLaMA-13B}   &  Self-Spec. & {pass@10}        & 0.659         & 1.345$\times$   \\
    \bottomrule
    \end{tabular}
    \caption{Evaluation on code generation tasks. We use greedy decoding for pass@1 and random sampling with a temperature of 0.6 for pass@10.}
    \label{tab:code}
\end{table}

We evaluate the performance of our decoding scheme, denoted as `Self-Speculative', with both greedy decoding (temperature = 0.0) and random sampling (temperature = 0.2/0.6), across text and code generation.
The baseline is `Autoregressive', which uses the original model to perform standard autoregressive decoding.
The experiments involve various scales of LLaMA-2 and its fine-tuned models. 
The results can be found in \Cref{tab:text,tab:code}.
We visualize the layer skipping distribution for different models in \Cref{apd:layers}.

For text generation tasks,  
\Cref{tab:text} shows that our method, when applied with temperature settings of 0.0 and 0.2 achieves considerable speedups ranging from 1.210$\times$ to \textbf{1.992$\times$}. Another important observation from these results is the minimal to nonexistent loss in ROUGE-2\footnote{
We attribute any slight differences observed in the case of greedy decoding to numerical rounding errors.
}, which verifies a core advantage of our decoding scheme, namely \textit{consistent output quality}. In \Cref{apd:quantitative}, we compare the ROUGE-2 with other mainstream LLMs and evaluate the various metrics (ROUGE-1 and ROUGE-L) to quantitatively show our output quality. Moreover, the case study in \Cref{apd:case_study} qualitatively presents consistent output examples.
In particular, our approach can be effectively applied on LLaMA-2-13B-Chat, a fine-tuned LLaMA-2-13B for conversation scenarios, indicating the compatibility of our method with fine-tuned models. 
This effectively addresses the dependency of the original speculative decoding method on high-quality draft models, which can be challenging to train and obtain, especially for fine-tuned models. 
Furthermore, the higher speedup achieved on LLaMA-2-70B suggests that larger models introduce more redundancy. This allows the drafting stage to skip a larger percentage of intermediate layers, thereby enhancing the overall speedup.

We also tested CodeLLaMA-13B, another fine-tuned LLaMA-2-13B for code generation. 
We used the HumanEval benchmark.
\Cref{tab:code} shows that we achieve speedups of 1.345$\times$ and 1.456$\times$, respectively, while maintaining similar task scores in terms of pass@1 and pass@10.
This further validates the compatibility of our scheme for coding.

\subsection{Impact of Skipped Layer Selection}

\begin{figure}[t]
    \centering
    \includegraphics[width=0.85\linewidth]{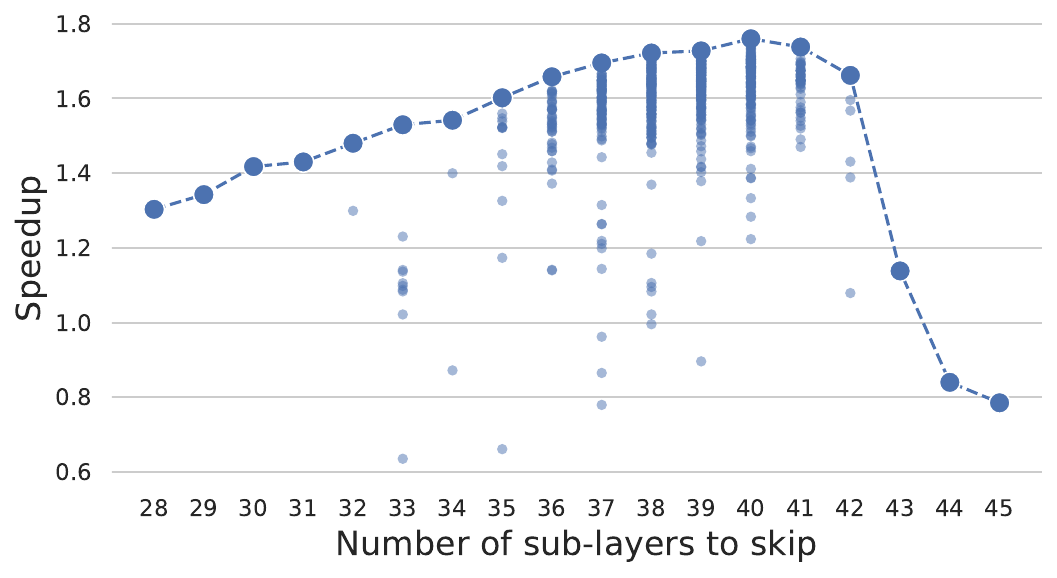}
    \caption{Speedup vs the number of skipped layers. These results are derived from the BO process.}
    \label{fig:layer_speedup}
\end{figure}

To investigate the impact of skipped layer selection, 
we conduct experiments on LLaMA-2-13B, which comprises 80 layers. 
Throughout the BO process, we track the number of layers skipped, denoted as $||\boldsymbol{z}^*||$, and the speedup relative to the autoregressive baseline.
\Cref{fig:layer_speedup} shows the results, where the dashed line indicates the maximum speedup for runs that skip the same number of layers.

These results reveal that:
(1) The peak end-to-end speedup is observed when about half of the layers are skipped during the drafting stage;
(2) The specific combination of layers skipped also plays a significant role. In particular, an inappropriate combination of skipped layers can actually result in a decrease in the end-to-end inference speed.
(3) There is a noticeable drop in speedup when more than 42 layers are skipped. This suggests that the quality of drafting significantly deteriorates when an excessive number of layers are omitted.

These findings indicate the importance of layer selection in the implementation of self-speculative decoding. However, alternative layer skipping strategies do not achieve satisfactory speedup compared to BO, as detailed in \Cref{apd:skipstra}. 

Performance degradation in drafting may be compensated by adopting \textit{aggressive skipping} strategy and further training the draft model on a small amount of data, as described in \Cref{apd:agg}. This finding aligns with the Sheared-LLaMA \cite{xia2023sheared}, which shows the effectiveness of pruning followed by fine-tuning on a small corpus.
 
\subsection{Effectiveness of Draft-Exiting}

Here we explore the effectiveness of the adaptive draft exit mechanism, specifically whether a threshold is needed and whether a static threshold is sufficient.
Our settings are LLaMA-2-13B, CNN/DM, and greedy decoding.

\textbf{Fixed Number of Draft Tokens.} 
We evaluated a self-speculative decoding variant where the draft model generates a constant number $K$ of tokens. 
As \Cref{tab:abla_K} shows, speedup initially increases with $K$, then decreases. 
This pattern arises as a large $K$ (e.g., $K=8$) produces many tokens likely to fail verification, wasting computation in drafting. 

While an appropriate $K$, e.g.~$K=4$, can partially alleviate this issue, a static setting limits the draft model's potential and achieves only a modest speedup (1.44$\times$). 
This is because a static $K$ does not adapt to the complexity of different instances. 
Ideally, we should use a larger $K$ for simpler instances and a smaller $K$ for challenging ones.

Moreover, \Cref{tab:abla_K} reveals that the acceptance rate and speedup do not have a direct proportionality. For instance, when $K=2$, the acceptance rate peaks at 0.924, yet the speedup is only 1.37$\times$. This discrepancy is due to the overly conservative $K=2$, which underutilizes the draft model's capacity. By generating fewer draft tokens, we miss opportunities to produce more valid draft tokens, thus limiting the overall speedup.

\textbf{Draft-Exiting with Static Threshold.} 
Another variant is to stop generating draft tokens if the confidence score falls below a predefined static threshold.
\Cref{tab:abla_gamma} shows that different static thresholds have large differences in acceleration (1.38$\times$\textasciitilde 1.58$\times$).
This highlights the importance of adaptively determining the appropriate threshold to optimize the speedup.
Specifically, we observe that high thresholds ($\gamma$=0.8) tend to underestimate the capabilities of the drafting model. Despite a high acceptance rate ($AR$=0.935), this does not necessarily result in the best speedup due to a reduced number of drafting tokens (1.55$\times$). Conversely, a lower threshold ($\gamma$=0.2) tends to overestimate the drafting model's capabilities, leading to a significantly lower acceptance rate ($AR$=0.749), wasting computational resources during the drafting stage, and thereby leading to slower inference speed (1.38$\times$).

\begin{table}[t]
    \centering
    \footnotesize
    \renewcommand{\arraystretch}{1.0}
    \setlength{\tabcolsep}{2.0mm}
    \begin{tabular}{@{}l|*{5}{c}@{}}
    \toprule
    $K$        & 2        & 4        & 6    & 8       & Adaptive    \\ 
    \midrule
    ROUGE-2    & 0.107   & 0.107     & 0.107   & 0.107  & \textbf{0.108}    \\
    $AR$         & \textbf{0.924}   & 0.865   & 0.807   & 0.748 & 0.919\\
    Speedup    & 1.37$\times$  & 1.44$\times$  & 1.42$\times$ & 1.36$\times$  & \textbf{1.57$\times$}   \\ 
    \bottomrule
    \end{tabular}
    \caption{Drafting with different $K$ values.
    }
    \label{tab:abla_K}
\end{table}

\begin{table}[t]
    \centering
    \footnotesize
    \renewcommand{\arraystretch}{1.0}
    \setlength{\tabcolsep}{2.0mm}
    \begin{tabular}{@{}l|*{5}{c}@{}}
    \toprule
    $\gamma$ & 0.2     & 0.4   & 0.6     & 0.8     & Adaptive \\
    \midrule
    ROUGE-2  & 0.107  & 0.107  & 0.107   & 0.107   & \textbf{0.108}    \\
    $AR$       & 0.749   & 0.852   & 0.909   & \textbf{0.935} & 0.919 \\
    Speedup  & 1.38$\times$ & 1.52$\times$ & \textbf{1.58}$\times$  & 1.55$\times$  & 1.57$\times$   \\ 
    \bottomrule
    \end{tabular}
    \caption{Static draft-exiting threshold $\gamma$ with $K=12$.}
    \label{tab:abla_gamma}
\end{table}
\textbf{Draft-Exiting with Adaptive Threshold.} 
To address the issue of optimal threshold determination, we propose an adaptive draft-exiting mechanism.
Specifically, we evaluate the acceptance rate and compare it to a target acceptance rate. The threshold is updated with an updating rule depicted in \Cref{sec:selection}.
\Cref{tab:abla_gamma} shows that the speedup achieved by our adaptive threshold update method (1.57$\times$) is comparable to, if not superior to, the speedup achieved with careful tuning of static thresholds. 
This indicates that dynamic threshold updating yields efficient and stable inference acceleration. This is mainly because the acceptance rate gets closer to the target $AR$ by adjusting the $\gamma$ value in a timely manner for instances of varying difficulties.
Also, \Cref{apd:maxtoken} reveals the adaptive draft-exiting is insensitive to changes in $K$.

\begin{figure}[t]
\centering
    \centering
    \includegraphics[width=0.85\linewidth]{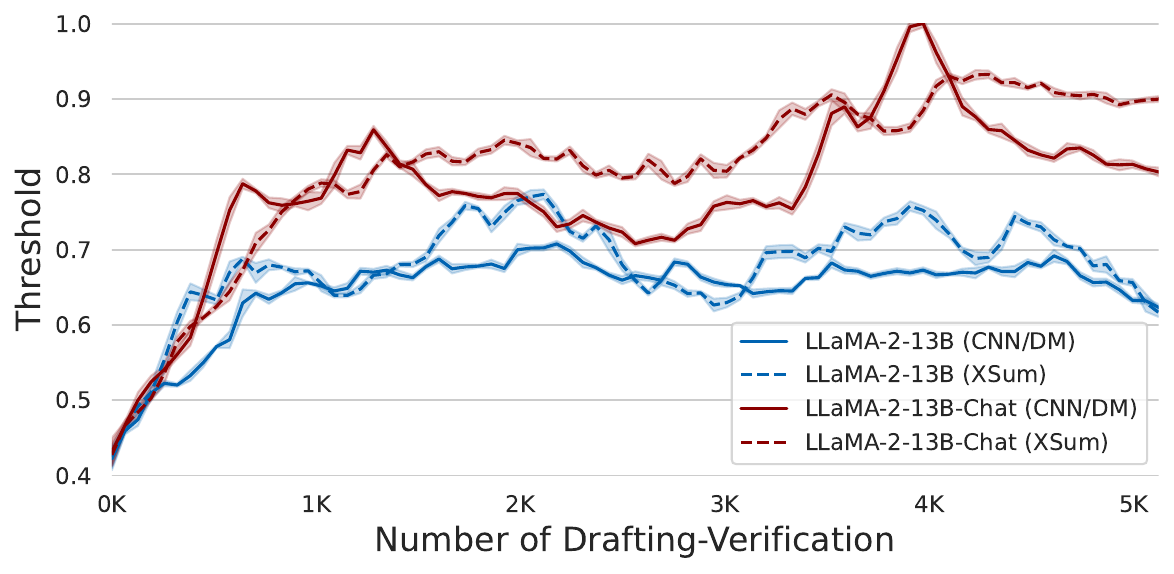}
    \caption{Threshold $\gamma$ varies with models and data. We calculate a moving average for every 64 cycles and plot the standard deviation. The initial $\gamma$ is set to 0.4.}
    \label{fig:threshold}
\end{figure}

\subsection{Evolution of Draft-Exiting Threshold}
\Cref{fig:threshold} captures the evolution of the draft-exiting threshold across different models and datasets over approximately 5,000 drafting and verification iterations. 
We can observe that our strategy adeptly adjusts the threshold, facilitating effective acceleration. However, the substantial variability across different models and datasets confirms the limitation of static threshold settings and the necessity for adaptive updates.

\subsection{Breakdown of Computation}
\begin{table}[ht]
    \centering
    \footnotesize
    \renewcommand{\arraystretch}{0.8}
    \begin{tabular}{lcc}
    \toprule
    Operation                      &  Autoregressive  & Self-Speculative  \\ 
    \midrule
    Drafting                       &  -               & \textbf{25.5$\pm$1.14 ms}  \\
    - Attention                    &  -               & 14.6$\pm$0.65 ms           \\
    - MLP                          &  -               & 9.46$\pm$0.42 ms           \\ 
    Verification                   &  -               & 10.7$\pm$2.81 ms           \\
    - Attention                    &  -               & 7.55$\pm$1.91 ms           \\
    - MLP                          &  -               & 2.73$\pm$0.80 ms           \\ 
    $\gamma$ update                &  -               & 0.61$\pm$0.14 \textmu s   \\ \midrule
    \rowcolor[HTML]{F2F2F2} 
    Latency *                        & 56.3$\pm$1.23 ms & 36.8$\pm$3.23 ms   \\
     
    - Attention                    & 39.7$\pm$0.91 ms      & 22.2$\pm$2.33 ms   \\
     
    - MLP                          & 14.3$\pm$0.29 ms      & 12.2$\pm$1.22 ms    \\   
    \bottomrule
    \end{tabular}
    \caption{Breakdown of computation. * denotes the average inference latency per token for 10 instances sampled from CNN/DM test set on LLaMA-2-13B.}
    \label{tab:breakdown}
\end{table}

\Cref{tab:breakdown} presents a computation breakdown comparing the baseline with our `Self-Speculative' decoding method. Our approach exhibits a significant speedup in average inference time per token compared to `Autoregressive'. This speedup is primarily attributed to two key techniques: the selection of skipped layers and the adaptive draft-exiting.
Notably, the drafting stage consumes the majority of inference latency, highlighting the need for draft model optimizations to improve overall inference speedup. Importantly, our adaptive exit mechanism ($\gamma$ update) incurs negligible computational cost as it does not involve neural network calculations.

\section{Conclusion}

In this paper, we introduced self-speculative decoding, a novel and efficient inference scheme that accelerates Transformer-based LLMs. 
Our method does not depend on additional neural network training and incurs no extra device memory, making it a highly practical and cost-effective solution for inference acceleration. 
Moreover, we used Bayesian optimization to search for layers to skip in drafting and proposed an adaptive draft-exiting mechanism to improve the end-to-end inference speed.
Benchmark tests with LLaMA-2 and its fine-tuned models demonstrated a speedup of up to 1.99$\times$.
For future work, we aim to explore more sophisticated model compression strategies to further accelerate the drafting stage for low-resource scenarios.

\section{Ethical Considerations}
In compliance with ethical considerations, we emphasize that the entirety of our research revolves around open-source datasets, models, and tools. Notably, we exclusively focus on improving model inference efficiency and do not engage in any commercial usage or ethical implications. 





\section{Limitations}
While our self-speculative decoding scheme presents benefits for accelerating LLMs, there are a few limitations to consider. 
Firstly, the utilization of Bayesian optimization to determine the layers to be skipped during the drafting stage may require several hours. Nonetheless, this limitation is not critical, as this process is a one-time, offline execution at the model level.
Secondly, our method does not involve any neural network training, which imposes a constraint on the number of layers that can be skipped. An excessive reduction in layers could result in a significant drop in the acceptance rate, thereby diminishing the achieved speedup.
Although fine-tuning the draft model--a sub-graph of the original model--could potentially mitigate this issue and yield a better speedup, as shown in \Cref{apd:agg}, it incurs additional memory overhead since the draft model no longer shares the same parameters with the original model.
In addition, we can refer to FlashAttention \cite{dao2023flashattention2} and vLLM \cite{kwon2023efficient} and similar works to further adapt our technique for batched decoding.

\bibliography{anthology,custom}
\bibliographystyle{acl_natbib}

\appendix

\section{Data} \label{apd:data}

The datasets that we have selected for evaluation are CNN/Daily Mail (CNN/DM), Extreme Summarization (XSum), and HumanEval. These tasks cover a broad spectrum of language processing capabilities, including text and code generation capabilities.
We perform 1-shot evaluation for CNN/DM and XSum, and compare the ROUGE-2.
We compare pass@10 for HumanEval.
For the results of efficiency, we randomly sample 1000 instances from the testset for CNN/DM and XSum.

\paragraph{CNN/Daily Mail (CNN/DM):} This task involves summarizing news articles from the CNN and Daily Mail websites. The models are required to generate a concise summary of each article, testing their ability to understand and condense complex information.

\paragraph{Extreme Summarization (XSum):} In the XSum task, models are asked to produce a single-sentence summary of a news article. This task tests the models' capability to extract the most salient information from a text and express it in a single, coherent sentence.

\paragraph{HumanEval:} The HumanEval task is a benchmark for Python programming. This task challenges the models with a variety of coding problems that require a wide range of skills, from basic programming to complex problem-solving abilities. It serves to evaluate the models' understanding of Python syntax, their ability to implement algorithms, and their proficiency in problem-solving using code. This benchmark provides a unique perspective on the models' capabilities in the realm of programming, complementing the language-focused tasks.

\section{Setup} \label{apd:setup}

We present the hyperparameter settings of the experiments in \Cref{tab:Hyper}, including the parameters involved in the decoding process, the adaptive draft-exiting mechanism, and the random sampling.
For the adaptive draft-exiting mechanism, we set the initial threshold $\gamma=0.6$, 
$\epsilon=0.01$, $\beta_{1}=0.5$, $\beta_{2}=0.9$, and $\alpha$ is slightly tuned for the data and model, as detailed in \Cref{tab:Hyper}.
For sampling-based inference, we by default use $top\_p=0.85$ for text summarization tasks, and $0.95$ for code generation tasks.

In addition, the key experimental environments on the A100-40GB are CUDA 11.6, PyTorch 1.13.1, and Transformer 4.33.1;
For the A100-80GB, the environment is CUDA 11.8, PyTorch 2.0.1, and Transformer 4.33.1.
We use an A100-40GB to conduct experiments for LLaMA-2-13B, LLaMA-2-13B-Chat, and CodeLLaMA-13B.
We use two A100-80GB with HuggingFace's accelerate\footnote{
\url{https://github.com/huggingface/accelerate} (Apache-2.0 License)} to conduct experiments for LLaMA-2-70B.

We randomly select 8 instances from the train set and use them to evaluate the inference time per token for Bayesian optimization.
This randomness not only ensures the generalizability of our approach but also mitigates any potential bias that could be introduced by the data selection process.
The offline Bayesian optimization time for 1000 iterations is about 2.5 hours for LLaMA-2-13B, LLaMA-2-13B-Chat, and CodeLLaMA-13B, and about 6 hours for LLaMA-2-70B.

\begin{table*}[t]
\centering
\small
\setlength{\tabcolsep}{2.6mm}
\begin{tabular}{@{}llccccccccc@{}}
\toprule
\multirow{2}{*}{Data} & \multirow{2}{*}{Model} & \multicolumn{2}{c}{Decoding} & \multicolumn{5}{c}{Adaptive Draft-Exiting} & \multicolumn{2}{c}{Random Sampling} \\ \cmidrule(l){3-4} \cmidrule(l){5-9} \cmidrule(l){10-11} 
                      &                       & $T$ & $K$ & $\alpha$ & $\epsilon$ & $\beta_{1}$ & $\beta_{2}$ & $\gamma_0$ & $top\_p$ & temperature \\ \midrule
CNN/DM                & LLaMA-2-13B           & 512 & 12  & 0.90     & 0.01       & 0.50        & 0.90        & 0.60       & 0.85     & 0.20       \\
CNN/DM                & LLaMA-2-13B-Chat      & 512 & 12  & 0.85     & 0.01       & 0.50        & 0.90        & 0.60       & 0.85     & 0.20       \\
CNN/DM                & LLaMA-2-70B           & 512 & 12  & 0.85     & 0.01       & 0.50        & 0.90        & 0.60       & 0.85     & 0.20       \\ \hline
XSum                  & LLaMA-2-13B           & 512 & 12  & 0.85     & 0.01       & 0.50        & 0.90        & 0.60       & 0.85     & 0.20       \\ 
XSum                  & LLaMA-2-13B-Chat      & 512 & 12  & 0.70     & 0.01       & 0.50        & 0.90        & 0.60       & 0.85     & 0.20       \\ 
XSum                  & LLaMA-2-70B           & 512 & 12  & 0.85     & 0.01       & 0.50        & 0.90        & 0.60       & 0.85     & 0.20       \\ \hline
HumanEval             & CodeLLaMA-13B         & 512 & 12  & 0.90     & 0.01       & 0.50        & 0.90        & 0.60       & 0.95     & 0.60        \\ \hline
GSM8K                 & CodeLLaMA-13B         & 512 & 12  & 0.90     & 0.01       & 0.50        & 0.90        & 0.60       & 0.95     & 0.60       \\
GSM8K                 & CodeLLaMA-13B-Instruct& 512 & 12  & 0.80     & 0.01       & 0.50        & 0.90        & 0.60       & 0.95     & 0.60   \\ \hline
MT-bench              & LLaMA-2-13B-Chat      & 512 & 12  & 0.85     & 0.01       & 0.50        & 0.90        & 0.60       & 0.85     & 0.20      
\\ \bottomrule
\end{tabular}
\caption{Hyperparameter settings. $\gamma_0$ represents the default initial value of $\gamma$.}
\label{tab:Hyper}
\end{table*}

\begin{figure*}[t]
    \centering

    \subfigure[LLaMA-2-13B]{
        \includegraphics[width=\textwidth]{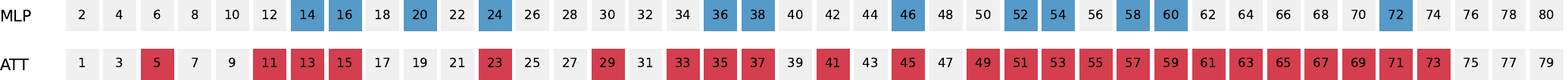}
        \label{fig:sub1}
    }
    
    \subfigure[LLaMA-2-13B-Chat]{
        \includegraphics[width=\textwidth]{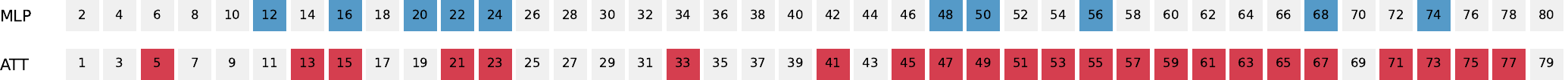}
        \label{fig:sub2}
    }

    \subfigure[CodeLLaMA-13B]{
        \includegraphics[width=\textwidth]{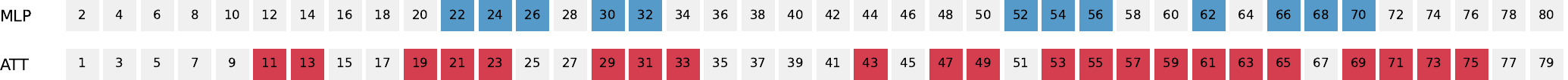}
        \label{fig:sub3}
    }

    \subfigure[The first half of LLaMA-2-70B]{
        \includegraphics[width=\textwidth]{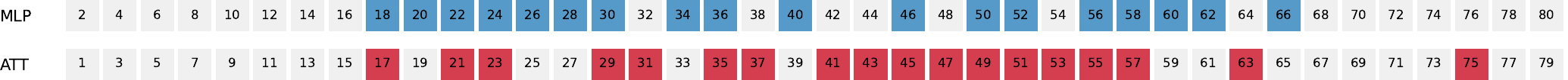}
        \label{fig:sub4}
    }

    \subfigure[The second half of LLaMA-2-70B]{
        \includegraphics[width=\textwidth]{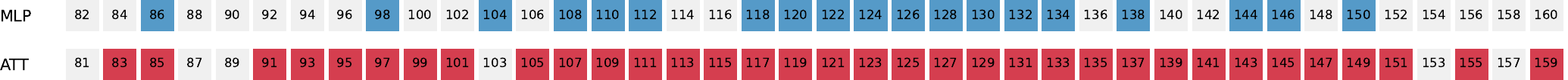}
        \label{fig:sub5}
    }

    \caption{Visualization of layer skip distributions in draft models for various base models. Gray squares indicate retained layers, red squares denote skipped attention layers, and blue squares signify skipped MLP layers.}
    \label{fig:skip}
\end{figure*}

\begin{table}[!ht]
\centering
\small
\setlength{\tabcolsep}{1.6mm}
\begin{tabular}{l|ccccc}
\toprule
Strategy    & First           & Mid.            & Last         & Rand.           & BO              \\ \midrule
ROUGE-2     & \textbf{0.108}  & \textbf{0.108}  & 0.107       & 0.107           & \textbf{0.108}        \\
$AR$          & 0.091          & 0.393           & 0.508       & 0.592           & \textbf{0.919}        \\
Speedup     & 0.696$\times$   & 0.887$\times$   & 0.951$\times$  & 1.01$\times$    & \textbf{1.57}$\times$ \\ \bottomrule
\end{tabular}
\caption{The effects of different skip strategies on the performance of CNN/DM on LLaMA-2-13B.}
\label{tab:skip_stra}
\end{table}

\begin{table}[!ht]
\centering
\small
\setlength{\tabcolsep}{2mm}
\begin{tabular}{l|ccccc}
\toprule
\#Iteration  & 200               & 400                & 600                     & 800               & 1000                     \\ \midrule
ROUGE-2     & \textbf{0.108}    & 0.107              & \textbf{0.108}          & \textbf{0.108}    & \textbf{0.108}            \\
$AR$          & 0.903             & \textbf{0.938}     & 0.920                   & 0.919             & 0.919           \\
Speedup     & 1.35$\times$      & 1.52$\times$       & \textbf{1.58}$\times$   & 1.57$\times$      & 1.57$\times$     \\ \bottomrule
\end{tabular}
\caption{The effect of the number of iterations of Bayesian optimization.}
\label{tab:iterBO}
\end{table} 

\begin{table}[!ht]
\centering
\small
\setlength{\tabcolsep}{1.8mm}
\begin{tabular}{l|ccccc}
\toprule
\#Token     & 0                & 20M        & 40M     & 60M      & 75M               \\ \midrule
ROUGE-2     & \textbf{0.106}            &  \textbf{0.106}       &  \textbf{0.106}   &   \textbf{0.106}      & \textbf{0.106}       \\
$AR$          & 0.695            &  0.902      &  0.901   &  0.900    & 0.893                 \\
Speedup     & 1.15$\times$    &   1.91$\times$      &  1.94$\times$   &  \textbf{1.96}$\times$    & 1.85$\times$ \\ \bottomrule
\end{tabular}
\caption{The effect of the number of tokens on aggressive skip performance.}
\label{tab:furthertrain}
\end{table} 

\begin{table}[!ht]
\centering
\small
\setlength{\tabcolsep}{2mm}
\begin{tabular}{l|ccccc}
\toprule
$K$         & 10                     & 12                     & 14                     & 16                    & 18                     \\ \midrule
ROUGE-2     & 0.107                  & \textbf{0.108}         & \textbf{0.108}         & \textbf{0.108}        & 0.107                   \\
$AR$          & \textbf{0.924}         & 0.919                  & 0.916                  & 0.913                 & 0.911                     \\
Speedup     & 1.57$\times$           & 1.57$\times$  & \textbf{1.58}$\times$  & \textbf{1.58}$\times$ & \textbf{1.58}$\times$     \\ \bottomrule
\end{tabular}
\caption{The effect of the max draft token under the adaptive draft-exiting mechanism.}
\label{tab:maxtoken}
\end{table}

\section{Which layers are skipped?} \label{apd:layers}

\Cref{fig:skip} visualizes the distinct base models corresponding to layer skip distributions within the draft model. Two key observations are made as follows:

First, we observe that there are more skips in the attention layer compared to the MLP layer, suggesting the attention layer is more effective for reducing inference time. This is reinforced by the results in \Cref{tab:breakdown}, where the time spent in the attention layer significantly contributes to the average inference latency per token.

Second, regardless of whether it is the MLP layer or the attention layer, the skipped layers tend to cluster in the latter half of the model. This pattern suggests that most tokens can be accurately predicted in the first half of the model, leaving the second half of the model relatively redundant.

\section{Effect of Skip Strategy} \label{apd:skipstra}
\begin{figure*}[ht]
    \centering

    \subfigure[First: Skip the first 24 layers of attention and 10 layers of MLP.]{
        \includegraphics[width=\textwidth]{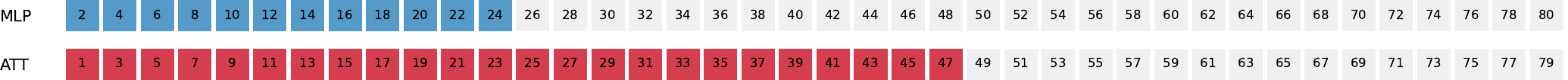}
        \label{fig:strasub1}
    }

    \subfigure[Middle: Skip the middle 24 layers of attention and 10 layers of MLP.]{
        \includegraphics[width=\textwidth]{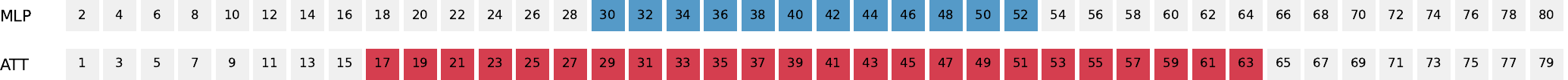}
        \label{fig:strasub2}
    }

    \subfigure[Last: Skip the last 24 layers of attention and 10 layers of MLP.]{
        \includegraphics[width=\textwidth]{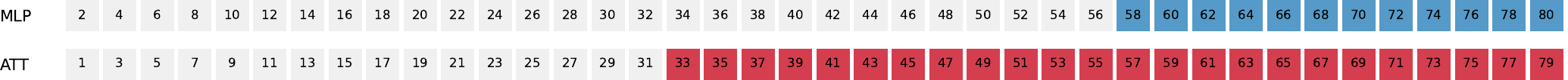}
        \label{fig:strasub3}
    }

    \subfigure[Random: Skip the 24 layers of attention and 10 layers of MLP randomly.]{
        \includegraphics[width=\textwidth]{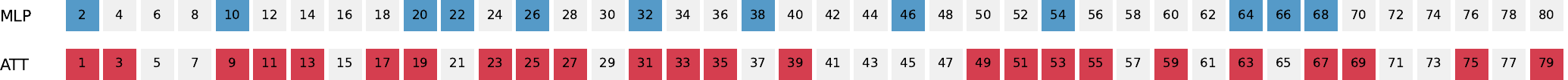}
        \label{fig:strasub4}
    }

    \subfigure[BO: Skip the 24 layers of attention and 10 layers of MLP by Bayesian Optimization.]{
        \includegraphics[width=\textwidth]{llama-2-13b-skip.pdf}
        \label{fig:strasub5}
    }

    \caption{Visualization of layer skip distributions in LLaMA-2-13B using different strategies.}
    \label{fig:straskip}
\end{figure*}

Here, we explore the effect of various skip strategies on the performance of generated draft models. 
We evaluate the CNN/DM on LLaMA-2-13B. 
Initially, we determine the number of skipped layers using Bayesian optimization, as illustrated in \Cref{fig:sub1}.
We find that the attention layer skips 24 layers, while the MLP layer skips 12 layers. 
We proceed to apply four strategies, each involving an equal number of layer skips: 
skipping the initial layers ("First"), middle layers ("Mid."), final layers ("Last"), and randomly sampling layers ("Rand.").
The layer skip distribution is visualized in \Cref{fig:straskip}.

\Cref{tab:skip_stra} reveals that the fixed strategies of layer skip (first, middle, last) or random skip yield minimal acceleration compared to Bayesian optimization (BO) results.
These suboptimal strategies, not optimized for average inference time, result in a draft model with subpar performance (manifested as very low $AR$), inefficient resource utilization in the drafting phase, and ultimately, a lack of speedup.
Furthermore, a slightly enhanced speedup is observed when skipping the last layers, likely due to the more severe redundancy in the model's final portion.

\section{Number of Iterations of BO} \label{apd:boiter}
\begin{figure*}[ht]
    \centering
    
    \subfigure[200 Iterations]{
        \includegraphics[width=\textwidth]{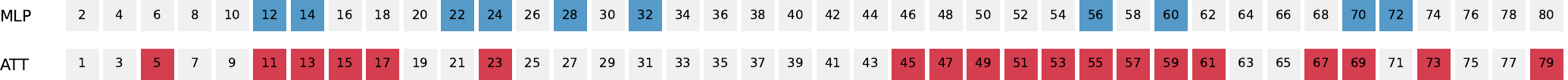}
        \label{fig:bosub1}
    }
    
    \subfigure[400 Iterations]{
        \includegraphics[width=\textwidth]{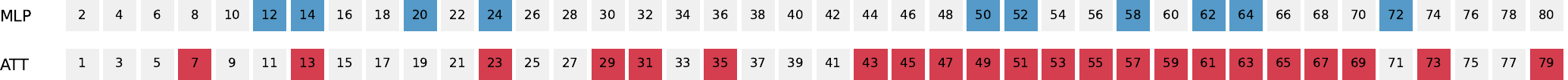}
        \label{fig:bosub2}
    }
    
    \subfigure[600 Iterations]{
        \includegraphics[width=\textwidth]{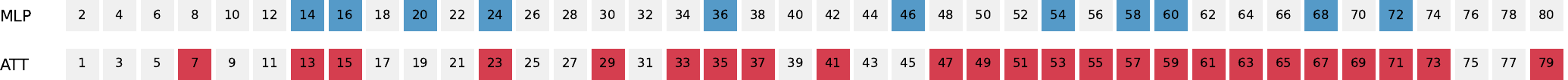}
        \label{fig:bosub3}
    }
    
    \subfigure[800 Iterations]{
        \includegraphics[width=\textwidth]{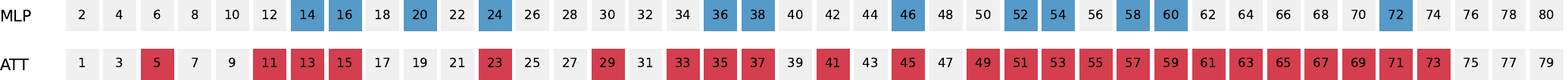}
        \label{fig:bosub4}
    }
    
    \subfigure[1000 Iterations]{
        \includegraphics[width=\textwidth]{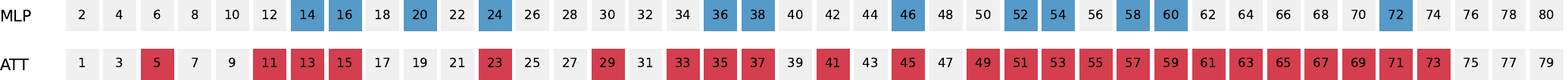}
        \label{fig:bosub5}
    }
    
    \caption{Visualization of LLaMA-2-13B layer skip distribution for different BO iteration numbers.}
    \label{fig:iterskip}
\end{figure*}

Subsequently, we explore the influence of the iteration number of Bayesian optimization on the performance of our decoding scheme and report the results in \Cref{tab:iterBO}. The layer skip distribution corresponding to different iteration numbers is depicted in \Cref{fig:iterskip}.

When applied to the LLaMA-2-13B for the CNN/DM task, we observe that while a higher number of iterations can yield increased acceleration, even a relatively modest number of iterations (e.g.,~200) effectively reduces inference time, achieving a speedup of 1.35$\times$. 
Notably, the performance metrics for the 800 and 1000 iterations exhibit consistency due to the same layers being skipped, as shown in \Cref{fig:iterskip}.

\section{Aggressive Skip} \label{apd:agg}
In pursuit of higher inference acceleration for users with ample resources, we explore a more aggressive skip strategy to obtain the draft model. 
To mitigate the performance degradation associated with the aggressive skip, we further train the draft model using 50,000 instances from the Pile dataset \cite{gao2020pile}, truncating the length of each to 512 tokens, which total up to 25 million tokens. We repeat this training for 3 epochs, resulting in a cumulative utilization of 75 million tokens.

\begin{figure*}[ht]
    \centering
    \includegraphics[width=\textwidth]{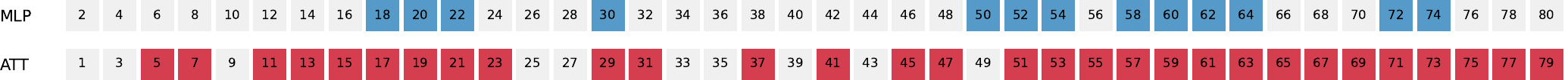}
    \caption{Visualize aggressive skip of 75\% attention layers and 32.5\% MLP layers of LLaMA-2-13B.}
    \label{fig:aggressiveskip}
\end{figure*}

The implementation of the aggressive skip involves skipping the top-K layers based on Bayesian optimization probabilities. For instance, in the case of LLaMA-2-13B, as illustrated in \Cref{fig:aggressiveskip}, we opt to skip 75\% of the attention layers (30 layers) and 32.5\% of the MLP layers (13 layers).

\Cref{tab:furthertrain} reveals that when we employ a more aggressive skip without further training (\#token is 0), there is a noticeable decrease in the draft model's quality, with an average acceptance rate of only 0.695.
This leads to a significantly reduced speedup of merely 1.15$\times$. 
Nevertheless, by dedicating a portion of the corpus to training, we notably enhance the draft model's quality, increasing the $AR$ to 0.900, in line with the target acceptance rate of 0.90. 
This enhancement enables a further improvement in speedup from 1.57$\times$ (shown in \Cref{tab:text}) to 1.96$\times$ (trained for 60M tokens), as more layers are skipped\footnote{
This finding aligns with the recent Sheared-LLaMA \cite{xia2023sheared}, which shows the effectiveness of pruning followed by further training on a small amount of data.}.
After training on 75 million tokens, the reason for the reduced acceleration is that we believe the model has a certain degree of overfitting.
It is essential to highlight that the aggressive skip strategy necessitates both an extended training process and the additional storage of trained draft models.
However, this trade-off is deemed acceptable for users with rich resources.


\section{Effect of Max \# of Draft Tokens} \label{apd:maxtoken}
Ideally, increasing the maximum number of draft tokens $K$ while maintaining a high acceptance rate should lead to further improvements in inference acceleration. To explore this, we test the CNN/DM task using LLaMA-2-13B, varying the max draft token $K$, and present the results in \Cref{tab:maxtoken}. 
It is noteworthy that as $K$ increases, the speedup remains relatively stable. This observation is primarily attributed to the fact that most tokens do not benefit from excessively large $K$ and tend to exit early.
In summary, our inference approach shows \textit{insensitivity} to $K$ thanks to the adaptive draft-exiting mechanism. 
Moreover, setting a relatively large value for $K$ (our default is 12) allows this mechanism to perform optimally.

\begin{table}[!t]
\centering
\small
\setlength{\tabcolsep}{2.6mm}
\begin{tabular}{l|cccccc}
\toprule
Quantization              & bf16           & fp8                   & fp4                & nf4 \\\midrule
ROUGE-2 $\uparrow$        & 0.107          & 0.105                 & 0.101              & \textbf{0.114} \\
$AR$      $\uparrow$        & 0.910          & 0.911                 & \textbf{0.913}     & 0.910 \\
VRAM (GB) $\downarrow$    & 37.2           & 27.5                  & \textbf{19.6}      & 19.7   \\
Latency (ms) $\downarrow$ & \textbf{32.4}  & 113                   & 152                & 126 \\
Speedup $\uparrow$        & 1.53$\times$   & \textbf{1.61}$\times$ & 1.36$\times$       & 1.35$\times$ \\
\bottomrule
\end{tabular}
\caption{Performance of self-speculative decoding combined with different quantization schemes of LLM.int8().}
\label{tab:quant}
\end{table}
\section{Adaptation} \label{apd:adaption}
\begin{figure*}[ht]
    \centering
    \includegraphics[width=\textwidth]{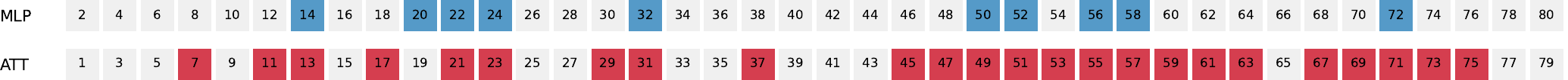}
    \caption{Visualization of layer skip distribution in  LLaMA-2-13B for quantization and sparsification.}
    \label{fig:adaption}
\end{figure*}

We here adopt skipping intermediate layers as a simple yet effective strategy to expedite the drafting stage. 
While other acceleration techniques such as quantization and structured pruning exist, they fail to offer speed-up proportional to their compression ratio.
Meanwhile, they require a separate copy of the altered model parameters, thereby increasing memory overhead. This contradicts the key requirement of no extra memory. Consequently, we adopt layer skipping in our approach. However, our scheme can be integrated with quantization  \cite{dettmers2022llm} and sparsification \cite{sun2023wanda} to further reduce resource consumption.
In this section, we explore the combination of self-speculative decoding with quantization and sparsification techniques to adapt to users with limited computing resources. We conduct experiments on the CNN/DM task using LLaMA-2-13B, and the layer skip distribution corresponding to the draft model is shown in \Cref{fig:adaption}.

\subsection{Quantization}
First, we integrate our inference approach with the quantization technique, LLM.int8()\footnote{\url{https://github.com/TimDettmers/bitsandbytes} (MIT License)} \cite{dettmers2022llm}. We evaluate the performance of three quantization schemes: fp8 (8-bit floating-point), fp4 (4-bit floating-point), and nf4 (4-bit normalized float), in comparison to the default bf16 (16-bit brain float point).
The results are presented in \Cref{tab:quant}.
In all quantization settings, we skip the `lm head' layer of the model and do not employ double quantization to save an additional 0.4 bits.

\Cref{tab:quant} illustrates that all three quantization schemes effectively reduce the video memory demand during inference. Notably, the fp4 quantization results in up to a nearly two-fold reduction in memory demand to just 19.6 GB. While there may be an increase in the average inference latency per token due to the dequantization process, this approach makes LLM suitable for scenarios with limited device memory.

\begin{table}[!t]
\centering
\small
\setlength{\tabcolsep}{1.8mm}
\begin{tabular}{l|cccc}
\toprule
Sparsification            & dense                   & unstructured          & 4:8                   & 2:4  \\ \midrule
ROUGE-2 $\uparrow$        & 0.107                   & 0.114                 & \textbf{0.115}        & 0.110 \\
$AR$      $\uparrow$        & 0.910                   & \textbf{0.918}        & 0.912                 & 0.911  \\
VRAM (GB) $\downarrow$    & 37.2                    & \textbf{35.9}         & \textbf{35.9}         & \textbf{35.9}    \\
Latency (ms) $\downarrow$ & 32.4                    & 30.4                  & \textbf{29.2}         & 30.9   \\
Speedup $\uparrow$        & \textbf{1.57}$\times$   & 1.50$\times$          & 1.47$\times$          & 1.48$\times$ \\
\bottomrule
\end{tabular}
\caption{Performance of self-speculative decoding combined with different sparsification schemes of wanda.}
\label{tab:spars}
\end{table}

\subsection{Sparsification}
Subsequently, we assess the performance of self-speculative decoding combined with sparsification techniques, specifically wanda\footnote{\url{https://github.com/locuslab/wanda} (MIT License)} \cite{sun2023wanda}, which includes unstructured sparsity and structured N:M sparsity (4:8 and 2:4) with the sparsity ratio of 0.5.
The N:M sparsity constraint specifies that no more than N out of every M contiguous weights can be non-zero.

\Cref{tab:spars} shows that while sparsification may not dramatically reduce VRAM requirements, it does result in a reduction in the average inference latency per token to varying degrees. 
However, the speedup is slightly down because the base model is also accelerated.

\begin{table}[!t]
    \centering
    \footnotesize
    \setlength{\tabcolsep}{0.2mm}
    \begin{tabular}{@{}lllcr@{}}
    \toprule
    Data       &Model         &Method       & Perfor. & Speedup \\ 
    \midrule
    {GSM8K}    &CodeLLaMA-13B & Autoreg.       & 0.104     &1.000$\times$\\
    {GSM8K}    &CodeLLaMA-13B & Self-Spec.      & 0.101     &1.351$\times$\\
    {GSM8K}    &CodeLLaMA-13B-Ins. & Autoreg.      & 0.223     &1.000$\times$\\
    {GSM8K}    &CodeLLaMA-13B-Ins. & Self-Spec.      & 0.213     &1.263$\times$\\
    {MT-bench} &LLaMA-2-13B-Chat & Autoreg.   & 6.940      &1.000$\times$\\
    {MT-bench} &LLaMA-2-13B-Chat & Self-Spec. & 6.930      &1.269$\times$\\
    \bottomrule
    \end{tabular}
    \caption{Evaluation on solving math problem and open-domain chitchat tasks. We use greedy decoding for the accuracy of GSM8K.
    We use random sampling for the average score at task-level of the MT-bench with various temperatures of 0.7 for `writing' and `roleplay', 0.1 for `stem' and `humanities', and greedy decoding for the rest of the tasks, according to \cite{zheng2023judging}.}
    \label{tab:diverse_task}
\end{table}
\section{Exploration of Diverse Tasks}
\label{apd:diverse_task}
We want to mention that our approach does not require model fine-tuning and its lossless characteristic is task-agnostic. However, we are open to evaluating the effectiveness of our approach on more diverse tasks.
Here, we further explore the performance of our decoding scheme on solving math problems (Grade School Math 8K \cite{cobbe2021gsm8k}) and open-domain chitchat (MT-bench \cite{zheng2023judging}).
For GSM8K, we evaluate on base model CodeLLaMA-13B and CodeLLaMA-13B-Instruct and report the problem-solving accuracy of 1318 questions in the test set. For MT-bench, our base model is LLaMA-2-13B-Chat, and we use its program script to ask GPT-4 to give a score of 10 points for each round using their prompt templates and report the average score for a total of 160 answers for two turns of 80 questions.

\paragraph{Grade School Math 8K (GSM8K):} A dataset of 8.5K high-quality linguistically diverse grade school math word problems. The dataset was created to support the task of question answering on basic mathematical problems that require multi-step reasoning. Solutions primarily involve performing a sequence of elementary calculations using basic arithmetic operations (+-×÷) to reach the final answer. This task is generally used to test logic and math in language modeling.

\paragraph{MT-bench:} A benchmark consisting of 80 high-quality multi-turn questions. MT-bench is designed to test multi-turn conversation and instruction-following ability, covering common use cases and focusing on challenging questions to differentiate models. The tasks identify 8 common categories
of user prompts to guide its construction: writing, roleplay, extraction, reasoning, math, coding, knowledge I (STEM), and knowledge II (humanities/social science). For each category, The task manually designed 10 multi-turn questions.

Regarding using Bayesian optimization 1,000 iterations to generate a draft model, for GSM8K, we randomly sampled 4 samples from the training set as a development set; for MT-bench, we let GPT-3.5 generate 8 examples according to 8 topics to form an optimized use development set.
Other relevant hyperparameters are shown in \Cref{tab:Hyper}.

For GSM8K, we achieve a 1.35$\times$ acceleration without compromising task performance. Regarding MT-bench, spanning writing, roleplay, reasoning, math, coding, extraction, stem, and humanities, presents substantial task diversity, posing challenges for generating high-quality and diverse output from draft models. Nevertheless, we still achieve a speedup of about 1.27$\times$ with lossless task performance.
These findings demonstrate the robustness and commendable performance of our approach across diverse tasks.

\section{Exploration of Output Quality} 
\subsection{Quantitative Analysis} \label{apd:quantitative}

\noindent
\textbf{LLMs Comparison.} Considering this non-fine-tuned, one-shot setting, our score is indeed quite competitive. We attach in \Cref{tab:other_LLMs} the performance of other mainstream large language models on CNN/DM and XSum as a reference (data from HELM leaderboard \cite{liang2022holistic}).

\begin{table}[t]
    \centering
    \small
    \setlength{\tabcolsep}{5mm}
    \begin{tabular}{@{}lcc@{}}
    \toprule
    Model & CNN/DM & XSum \\
    \midrule
    LLaMA-2-13B (ours)  & 0.106 & 0.124 \\
    LLaMA-2-13B-Chat (ours) & 0.144 & 0.109 \\
    LLaMA-2-70B (ours) & 0.130 & 0.118 \\
    OPT (66B) & 0.136 & 0.126 \\
    Davinci (175B) & 0.127 & 0.126 \\
    Palmyra X (43B) & 0.049 & 0.149 \\
    GPT-NeoX (20B) & 0.123 & 0.102 \\
    Luminous Base (13B) & 0.110 & 0.105 \\
    BLOOM (176B) & 0.080 & 0.030 \\
    YaLM (100B) & 0.017 & 0.021 \\
    \bottomrule
    \end{tabular}
    \caption{Our model with self-speculative decoding vs. Mainstream LLMs with autoregressive.}
    \label{tab:other_LLMs}
\end{table}

\begin{table}[!t]
    \centering
    \small
    \setlength{\tabcolsep}{1.8mm}
    \begin{tabular}{@{}llcccr@{}}
    \toprule
    \multirow{2}{*}{Model}
    & \multirow{2}{*}{Method}
    & \multicolumn{3}{c}{ROUGE} 
    & \multirow{2}{*}{Speedup} \\
    \cmidrule(l){3-5}
           &      & 1 &2    &L &  \\ 
    \midrule
    {CNN/DM}   &  Autoreg.   &0.263    &0.106    &0.181   & 1.000$\times$\\
    {CNN/DM}   &  Self-Spec. &0.266    &0.108    &0.183   & 1.429$\times$\\
    {XSum}     &  Autoreg.   &0.327    &0.124    &0.260   & 1.000$\times$\\
    {XSum}     &  Self-Spec. &0.328    &0.125    &0.261   & 1.377$\times$\\
    \bottomrule
    \end{tabular}
    \caption{Evaluation on text generation tasks with various metrics and greedy decoding.}
    \label{tab:diverse_metric}
\end{table}

\noindent
\textbf{Diverse Metrics.} Our experimental setup mainly follows the work of DeepMind \cite{chen2023accelerating}. For summarization tasks, ROUGE-2 is a widely recognized evaluation metric, as adopted by \cite{chen2023accelerating,liang2022holistic}, in addition to ROUGE-1 and ROUGE-L. However, we want to emphasize that we have not disregarded the importance of other metrics such as ROUGE-1 and ROUGE-L. Indeed, we have maintained records of these results as part of our comprehensive analysis, and we will attach them in the appendix in the revision. The results are presented below. And we can see that our conclusions remain consistent – we have been successful in achieving a significant acceleration in task performance without compromising the quality of the outputs. We believe this reaffirms the robustness and efficacy of our approach.

\subsection{Qualitative Analysis} \label{apd:case_study}
We present more qualitative evidence to supplement the quantitative ROUGE-2 score. Please note, however, that our approach is inherently lossless and the performance evaluation is conducted on the original model without any fine-tuning. The task performance of our self-speculative decoding is therefore reliant on the inherent capabilities of the base model utilizing autoregressive decoding. Specifically, using the same base model, in the greedy setting, generation results from self-speculative decoding are \textbf{identical} to those of autoregressive decoding; In the sampling setting, the results generated are from the same distribution as those for autoregressive decoding. These properties of speculative decoding have been mathematically proven in previous work \cite{leviathan2023fast}. Finally, to address concerns, we present a case study on LLaMA-2-13B, detailed in the \Cref{tab:case_study}. 

\section{Algorithm with Sampling} \label{apd:samping_algo}

\begin{table*}[!t]
    \centering
    \small
    \begin{tabular}{@{}llp{3cm}p{3cm}p{3cm}p{3cm}@{}}
    \toprule
    \multirow{2}{*}{Data}
    & \multirow{2}{*}{Temp.}
    & \multirow{2}{*}{Article/Document}
    & \multicolumn{3}{c}{Summary/Highlight} \\
    \cmidrule(l){4-6}
           &      &    &Reference &Autoregressive  &Self-Speculative  \\ 
    \midrule
    {CNN/DM}   &  0.0   
    &An Arkansas woman became a little richer this week after finding a large diamond at Crater of Diamonds State Park. Susie Clark, of Evening Shade, found the 3.69-carat white diamond in a 37.5-acre search field on Thursday, her last day of a trip to the park with her husband.   Clark,
    &Susie Clark, of Evening Shade, Arkansas, found the diamond on Thursday at Crater of Diamonds State Park . The 3.69-carat gem is teardrop-shaped about the size of a pinto bean . Park interpreter said it's the largest of the 122 diamonds found this year .    &woman who found a 3.69-carat white diamond at Crater of Diamonds State Park in Arkansas has named the gem a 'hallelujah diamond' because she spotted the gem shortly after praying that she would find something . 
    & woman who found a 3.69-carat white diamond at Crater of Diamonds State Park in Arkansas has named the gem a 'hallelujah diamond' because she spotted the gem shortly after praying that she would find something .\\
    {CNN/DM}   
    &  0.2  
    & who first visited the park 33 years ago with her mother and grandmother, prayed to God, asking, 'Are you doing to bless me and let me find a diamond today?' as she searched the field. Susie Clark, of Evening Shade, Arkansas, found a 3.69-carat white ...   
    &Susie Clark, of Evening Shade, Arkansas, found the diamond on Thursday at Crater of Diamonds State Park . The 3.69-carat gem is teardrop-shaped about the size of a pinto bean . Park interpreter said it's the largest of the 122 diamonds found this year .    &woman who found a 3.69-carat diamond at Crater of Diamonds State Park in Arkansas has named the gem a 'hallelujah diamond' because she spotted the gem shortly after praying that she would find something .   
    &woman who found a 3.69-carat white diamond at Crater of Diamonds State Park in Arkansas has named the gem a 'hallelujah diamond' because she spotted the gem shortly after praying that she would find something .\\
    {XSum}     &  0.0   
    &The flight provider operates a "bug bounty" scheme that rewards hackers for privately disclosing security flaws rather than sharing them     
    & US airline United has rewarded two hackers who spotted security holes in its website with a million free flight miles each.     
    & Airlines has given out its first rewards to hackers who reported security flaws in its website.   
    & Airlines has given out its first rewards to hackers who reported security flaws in its website.\\
    {XSum}     &  0.2  
    & online. It has given the maximum reward of a million flight miles, worth dozens of trips, to two people. One security expert said the ...     
    & US airline United has rewarded two hackers who spotted security holes in its website with a million free flight miles each.        
    & airline has given out two million flight miles to hackers who found security flaws in its website.   
    & airline has given the maximum reward of a million flight miles, worth dozens of trips, to two people.\\
    \bottomrule
    \end{tabular}
    \caption{Case study of CNN/DM and XSum on LLaMA-2-13B.}
    \label{tab:case_study}
\end{table*}

We first demonstrate self-speculative sampling with greedy sampling by integrating the adaptive draft-exiting mechanism in \Cref{alg:SSD_complete}.

In addition to the greedy version of self-speculative decoding that we have presented in the main paper, we also explore a variant that incorporates random sampling, that also incorporated a complete adaptive draft-exiting mechanism, as shown in \Cref{alg:full_SSD}.
This approach introduces an element of randomness into the selection of tokens for speculative decoding, as opposed to the deterministic nature of the greedy version.
In our setup, random sampling is affected by two parameters: temperature and $top_p$. Higher values of temperature or $top_p$ lead to greater token diversity, while lower values make token selection more deterministic.
This variant could potentially lead to diverse decoding paths and outcomes, which may be beneficial in certain scenarios, such as code generation.

\begin{algorithm}[ht]
\small
\caption{Self-Speculative Decoding (Greedy) }\label{alg:SSD_complete}
\begin{algorithmic}[1]
\State LLM $p(x|\boldsymbol z^*,x_1, ..., x_t)$ where $x_1, ..., x_t$ is the prompt, $\boldsymbol z^*$ is a vector that represents the specific layers to bypass;
target sequence length $T$;
max draft tokens to generate $K$.
We denote the original LLM as $p(x|\vec{0},x_1, ..., x_t)$, where $\vec{0}$ is a zero vector, indicating all layers are used in inference.
We denote the acceptance rate ($AR$) at $e$-th drafting stage as $AR_e$. 
\State $i \leftarrow t$
\While{$i < T$}
\For{$j \leftarrow i, ..., i + K$} \Comment{Drafting Stage}
    \State $x_{j+1} \leftarrow \argmax{p(x|\boldsymbol z^*,x_1, ..., x_{j})}$    
    \If{$\max{p(x|\boldsymbol z^*,x_1, ..., x_{j})} < \gamma$} \Comment{Draft Exit}
        \State Break
    \EndIf
\EndFor
\For{$i \leftarrow i, ..., j$} \Comment{Verification Stage}
    \If{$x_{i+1} \ne \argmax{p(x|\vec{0}, x_1, ..., x_{i})}$}
        \State $x_{i+1} \leftarrow \argmax{p(x|\vec{0}, x_1, ..., x_{i})}$ 
        \State Break
    \EndIf
\EndFor
\State $i \leftarrow i + 1$
\State If all draft tokens are accepted, 
        generate next token $x_{i+1} \leftarrow \argmax{p(x|\vec{0},x_1, ..., x_{i})}$ and $i \leftarrow i+1$
\State $AR \gets \beta_{1} AR + (1 - \beta_{1})AR_e$ \Comment{$\gamma$ Update }
\If{${AR} \leq \alpha$}
    \State $\gamma + \epsilon$
\Else
    \State $\gamma - \epsilon$
\EndIf
$\gamma \gets \beta_{2} \gamma + (1 - \beta_{2}) \tilde{\gamma}$
\EndWhile
\State \textbf{return} $x_1, ..., x_T$
\end{algorithmic}
\end{algorithm}

\begin{algorithm*}[t]
\centering
\small
\caption{Self-Speculative Decoding}\label{alg:full_SSD}
\begin{algorithmic}[1]
\State LLM $p(x|\boldsymbol z^*,x_1, ..., x_t)$ where $x_1, ..., x_t$ is the prompt, $\boldsymbol z^*$ is a vector that represents the specific layers to bypass;
target sequence length $T$;
max draft tokens to generate $K$.
We denote the original LLM as $p(x|\vec{0},x_1, ..., x_t)$, where $\vec{0}$ is a zero vector, indicating all layers are used in inference.
We denote the acceptance rate ($AR$) at $e$-th drafting stage as $AR_e$.
\State $i \leftarrow t$
\While{$i < T$}
\For{$j \leftarrow i, ..., i + K$} \Comment{Drafting Stage}
    \State $x_{j+1} \leftarrow \sample{p(x|\boldsymbol z^*,x_1, ..., x_{j})}$    
    \If{$\max{p(x|\boldsymbol z^*,x_1, ..., x_{j})} < \gamma$} \Comment{Draft Exiting}
        \State Break
    \EndIf
\EndFor
\For{$i \leftarrow i, ..., j$} \Comment{Verification Stage}
    \State $r \leftarrow \sample$ from a uniform distribution $U[0,1]$
    \If{$r \geq \min(1, \frac{p(x|\vec{0}, x_1, ..., x_{i})}{p(x|\boldsymbol z^*,x_1, ..., x_{i})})$}
        \State $x_{i+1} \leftarrow \sample$ from $\frac{\max(0,p(x|\vec{0}, x_1, ..., x_{i})-p(x|\boldsymbol z^*,x_1, ..., x_{i}))}
        {\sum_{x}{\max(0,p(x|\vec{0}, x_1, ..., x_{i})-p(x|\boldsymbol z^*,x_1, ..., x_{i}))}}$
        \State Break
    \EndIf
\EndFor
\State $i \leftarrow i + 1$
\State If all draft tokens are accepted, 
        generate next token $x_{i+1} \leftarrow \sample{p(x|\vec{0},x_1, ..., x_{i})}$ and $i \leftarrow i+1$
\State $AR \gets \beta_{1} AR + (1 - \beta_{1})AR_e$ \Comment{$\gamma$ Updating }
\If{${AR} \leq \alpha$}
    \State $\gamma + \epsilon$
\Else
    \State $\gamma - \epsilon$
\EndIf
$\gamma \gets \beta_{2} \gamma + (1 - \beta_{2}) \tilde{\gamma}$
\EndWhile
\State \textbf{return} $x_1, ..., x_T$
\end{algorithmic}
\end{algorithm*}

\clearpage

\end{document}